\documentclass[runningheads]{llncs}

\usepackage{eccv}

\usepackage{eccvabbrv}

\usepackage{graphicx}
\usepackage{booktabs}

\usepackage{multirow}

\usepackage{booktabs,tabularx}

\usepackage[table]{xcolor}

\usepackage[accsupp]{axessibility}  % Improves PDF readability for those with disabilities.

\usepackage{hyperref}

\usepackage{orcidlink}
\usepackage{enumerate}
\usepackage{enumitem}
\usepackage{marvosym}

\usepackage{wrapfig}

\usepackage{booktabs}
\usepackage{tabularx}
\usepackage{xcolor,colortbl}
\usepackage{makecell}

\begin{document}

\title{MA-VLA: Multi-Arm Vision-Language-Action Model for Collaboration and Compositional Generalization}

\titlerunning{MA-VLA for Collaboration and Compositional Generalization}

\author{
Zaibin Zhang$^{1*}$,
Junlan Xiao$^{1*}$,
Zhongbo Zhang$^{1*}$,
Yifan Wang$^{1}$,
Li Kang$^{4}$,
Yiran Qin$^{6\dagger}$,
Changxing Xia$^{1}$,
Heng Zhou$^{5}$,
Talas Fu$^{1}$,
Enshen Zhou$^{7}$,\\
Ruimao Zhang$^{3}$,
Zhenfei Yin$^{2\dagger}$,
Huchuan Lu$^{1}$,
Lijun Wang$^{1\dagger}$\textsuperscript{\Letter}
}

\authorrunning{Z.~Zhang et al.}

\institute{
$^{1}$ Dalian University of Technology,
$^{2}$ University of Oxford,
$^{3}$ Sun Yat-sen University\\
$^{4}$ Shanghai Jiao Tong University,
$^{5}$ University of Science and Technology of China\\
$^{6}$ The Chinese University of Hong Kong, Shenzhen,
$^{7}$ Beihang University\\[2pt]
\texttt{dlutzzb@gmail.com},
\texttt{ljwang@dlut.edu.cn}\\[2pt]
$^*$ Equal contribution
\qquad
$^\dagger$ Equal advising
\qquad
\textsuperscript{\Letter} Corresponding author
}

% Second Author\inst{2,3}\orcidlink{1111-2222-3333-4444} \and
% Third Author\inst{3}\orcidlink{2222--3333-4444-5555}}

% % TODO FINAL: Replace with an abbreviated list of authors.

% % TODO FINAL: Replace with your institution list.
% Springer Heidelberg, Tiergartenstr.~17, 69121 Heidelberg, Germany

\maketitle
\begin{abstract}
Multi-arm collaboration is becoming a core capability in embodied manipulation. Recent vision-language-action (VLA) models integrate perception, language, and control, but most represent language as a single global instruction and do not provide an explicit mechanism for assigning and composing arm-specific behaviors. This design limits transfer to collaboration patterns that differ from those observed during training.
We present MA-VLA, a unified framework for multi-arm collaboration via atomic action assignment. MA-VLA decomposes cooperative behavior into mid-level atomic prompts and allocates them to individual arms, enabling explicit subgoal specification and compositional reuse across tasks. To reduce reliance on fixed execution roles, we introduce Arm Shuffle, a training-time permutation of the observation, state, and assigned atomic prompts for each arm. This permutation enforces role-agnostic instruction following and supports recomposition into unseen coordination patterns, which we term multi-arm compositional generalization. We also construct a benchmark in which test-time collaboration patterns are absent in training set. Across simulation and real-world evaluations, prior state-of-the-art VLAs largely fail under these unseen collaborations, while MA-VLA consistently succeeds. These results indicate that structured, per-arm atomic action assignment offers a practical route to scalable generalization in multi-arm embodied systems. Code, models, and data are available at \url{https://github.com/zhangzaibin/future-robots}

\keywords{Multi-arm collaboration \and Compositional generalization \and Vision-language-action}
\end{abstract}
    
\section{Introduction}

Embodied Artificial Intelligence (Embodied AI) studies how intelligent systems perceive, reason, and act through continuous interaction with the physical world~\cite{duan2022survey}.
In robotics, this interaction becomes substantially more complex when the setting shifts from single-arm manipulation to multi-arm collaboration~\cite{lai2025roboballet,black2024pi_0,mu2024robotwin,qin2025robofactory,team2025gemini,cheang2025gr,jiang2025rethinking, im2025twinvla,lu2025anybimanual}.
By leveraging parallel execution and coordinated control, multi-arm and multi-robot systems can solve tasks that are infeasible for a single arm, which makes collaboration a central direction in embodied manipulation.

\begin{figure}[t]
\centering
\includegraphics[width=1.0\linewidth]{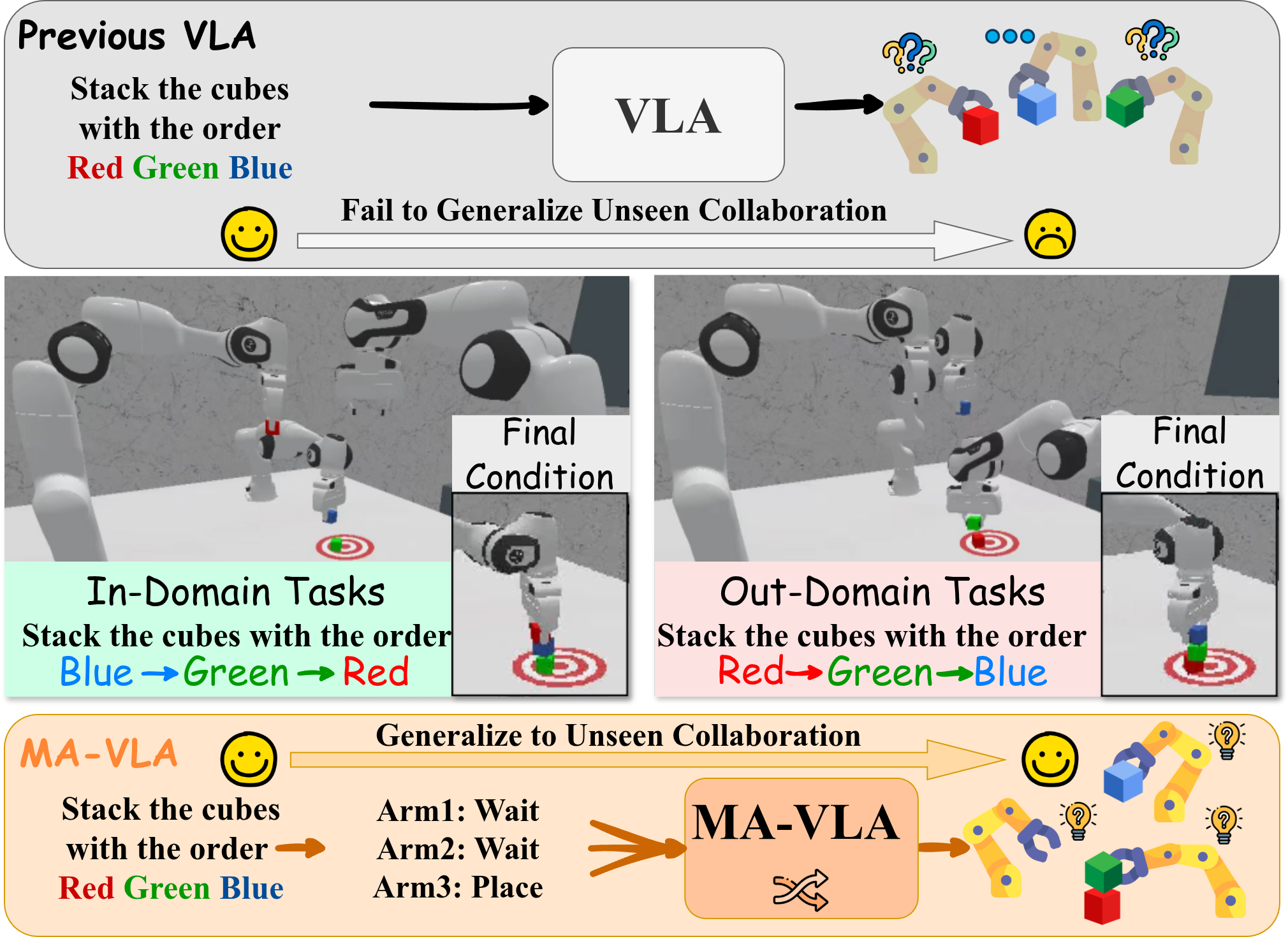}
\caption{Comparison with prior VLA systems. Previous VLAs typically execute a task from a single high-level instruction. MA-VLA instead decomposes the instruction into a sequence of interpretable mid-level atomic actions and assigns them to individual arms, which enables recomposition to handle unseen collaboration patterns.}
\label{fig:teaser}
\end{figure}

Recent progress has trained dual-arm and multi-arm systems with imitation learning~\cite{qin2025robofactory,mu2024robotwin,jiang2025rethinking,lu2025anybimanual,lee2024interact} or reinforcement learning~\cite{lai2025roboballet}.
In parallel, end-to-end vision-language-action (VLA) models~\cite{ma2024vlasurvey} have enabled dual-arm robots to follow natural-language instructions for manipulation~\cite{black2024pi_0,pertsch2025fast,intelligence2025pi_,chen2025robotwin,team2025gemini,cheang2025gr}.
Despite these advances, most existing VLAs still treat language as a single static command that is mapped to control, without an explicit mechanism for task decomposition and sub-task allocation across arms.
This design choice makes collaboration brittle, because division of labor must be inferred implicitly from data and often becomes over-specialized to a small set of training collaboration patterns.

In human teamwork, effective cooperation is driven by division of labor~\cite{minsky1986society}: complex goals are achieved by assigning mid-level responsibilities and executing atomic actions that compose into a coherent plan.
This structure is also a natural source of generalization, because new strategies can be formed by recombining familiar atomic actions with familiar assignment patterns.
Motivated by this observation, we study a setting that is largely missing in current multi-arm VLA research: whether a learned system can handle unseen collaboration patterns by recomposing known atomic actions.
We refer to this capability as \textbf{multi-arm compositional generalization}. However, compositional generalization is non-trivial: test-time coordination often requires unseen combinations of layouts, arm states, and inter-arm dependencies, which cannot be addressed by role swapping or independently trained arms.

To address this gap, we propose MA-VLA, a unified vision-language-action framework for multi-arm collaboration via atomic action assignment.
Given a high-level instruction, MA-VLA outputs a sequence of interpretable mid-level atomic actions and assigns them to each arm within a shared model.
This representation makes division of labor explicit and offers a simple interface for coordinating execution across arms, while remaining end-to-end at inference time.

To reduce reliance on fixed arm identities and improve compositional transfer, we introduce Arm Shuffle, a training-time permutation that randomizes the mapping between arms and their input bundles, including arm state, local observation, and assigned atomic prompts.
By preventing the model from binding behaviors to arm indices, Arm Shuffle promotes role-agnostic instruction following and enables recomposition of atomic actions into unseen collaboration patterns, such as novel stacking orders and object-passing strategies.

We evaluate MA-VLA on RoboFactory~\cite{qin2025robofactory}, RoboTwin 2.0~\cite{chen2025robotwin}, and a real-world dual-arm SO101 platform under both in-domain settings and multi-arm compositional out-of-domain splits. MA-VLA consistently outperforms competitive imitation and VLA baselines, with the largest gains observed in both in-domain performance and compositional generalization, suggesting a practical path toward flexible and scalable multi-arm collaboration.

\section{Related Work}

\subsection{Multi-arm Robot Manipulation}
Behavior Cloning (BC)~\cite{dalal2023imitating,osa2018algorithmic,o2024open,jang2022bc,jiang2022vima,mandlekar2020learning,kim2024openvla} and Offline Reinforcement Learning (ORL)~\cite{kalashnikov2021mt,chebotar2023q,kumar2022pre} remain dominant paradigms for robotic policy learning but rely heavily on large-scale expert demonstrations. Reinforcement learning based imitation~\cite{ho2016generative,papagiannis2022imitation} mitigates this dependency yet demands costly online exploration. Recently, generative methods such as Action Chunking Transformer (ACT)~\cite{buamanee2024bi,vaswani2017attention,zhao2023learning} and Diffusion Policy~\cite{chi2025diffusion,jiang2025rethinking,ze20243d}, empowered by scalable simulation frameworks~\cite{james2020rlbench,nasiriany2024robocasa,qin2024worldsimbench,tao2024maniskill3,mu2024robotwin,zhu2020robosuite}, have shown strong modeling and generalization ability. Most studies~\cite{black2024pi_0,mu2024robotwin, team2025gemini, cheang2025gr, jiang2025rethinking, shi2025hi} still focus on bimanual tasks, while RoboBallet~\cite{lai2025roboballet}  achieves coordination among three or more arms via reinforcement learning, RoboFactory~\cite{qin2025robofactory} leverages Diffusion Policy for multi-arm collaboration, while research on multi-arm via VLA remains largely unexplored.

\subsection{Vision-Language-Action Models}
Vision-Language-Action (VLA) models~\cite{zitkovich2023rt,kim2024openvla,team2024octo,black2024pi_0,team2025gemini,zhang2025dreamvla,qi2025sofar,zhang20254d,qu2025spatialvla,zhou2025vision,wang2025vq,li2025bridgevla,driess2025knowledge,wen2025dexvla,li2025hamster,huang2025otter,liu2025hybridvla,wu2025momanipvla,zhao2025cot} bridge perception and control by integrating vision, language, and action generation. With advances in LLMs~\cite{touvron2023llama,brown2020language,achiam2023gpt} and VLMs~\cite{liu2023visual,karamcheti2024prismatic,li2022blip, zhou2026roborefer, team2025robobrain, zhou2025robotracer, tan2026robobrain,yu2025far}, they leverage cross-modal reasoning to map visual inputs and natural instructions into robot actions.
Early systems like RT2~\cite{zitkovich2023rt} and Octo~\cite{team2024octo} learn from large-scale multimodal data, while OpenVLA~\cite{kim2024openvla} extends pretraining to the Open X-Embodiment dataset~\cite{o2024open}. pi0~\cite{black2024pi_0} further introduces flow-matching for continuous control. Yet, these works treat language as static instruction rather than an interactive medium, limiting generalization. MAVLA employs language as an active coordination channel, enabling compositional multi-arm behaviors.

\subsection{Task Generalization}
Prior studies~\cite{yenamandra2023homerobot,pumacay2024colosseum,gupta2019relay,hua2024gensim2,chen2025robotwin,li2025hamster} primarily address generalization to environmental or visual changes, such as object variations~\cite{mu2021maniskill,li2024evaluating,akinola2025tacsl}. Recent work extends to task-level generalization through compositional~\cite{yang2025embodiedbench,yu2020meta,liu2023libero,wang2025roboeval} and parameterized policies~\cite{mees2022calvin,gong2023arnold,garcia2025towards}.
Nonetheless, these methods largely focus on single- or dual-arm setups. Our \textit{Arm Shuffle} tackles multi-arm collaborative generalization by randomizing arms’ roles during training, breaking fixed dependencies and promoting flexible coordination.

\section{Preliminaries}

\begin{figure*}[t]
    \centering
    \includegraphics[width=1.0\linewidth]{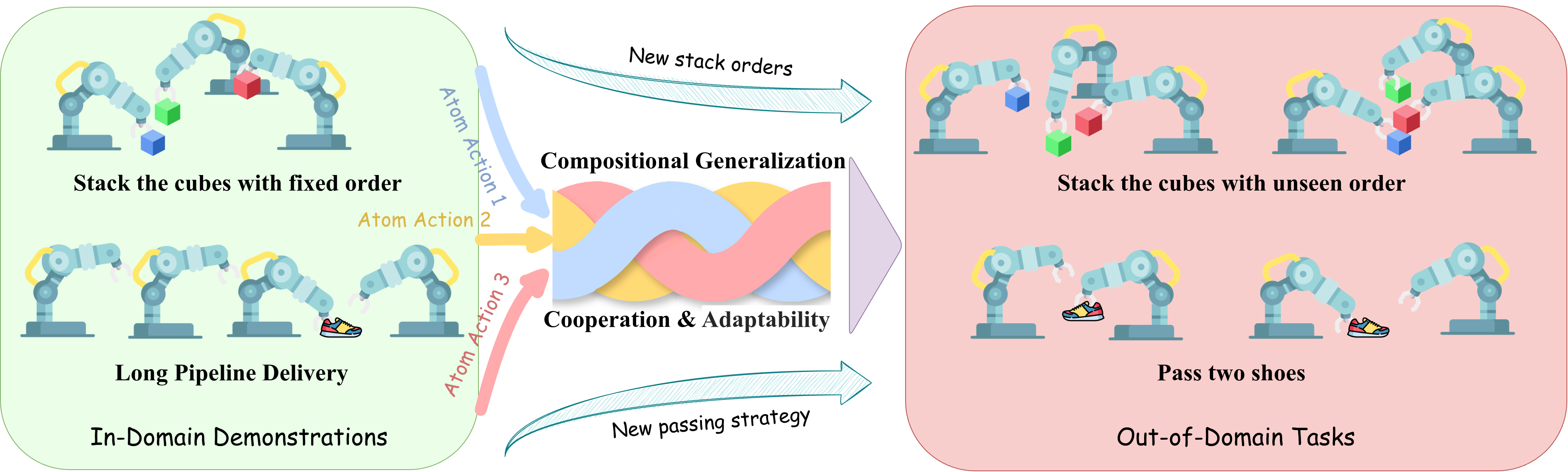}
    \caption{Multi-arm compositional generalization: unseen collaboration patterns induced by new combinations of arm roles, spatial states, and atomic actions, beyond the patterns observed in training.}
    \label{fig:compositional_generalization}
\end{figure*}

\subsection{Vision-Language-Action Models for Multi-arm Manipulation}

\paragraph{Single-arm VLA.}
At timestep $t$, the robot observes $\mathbf{o}_t=\{\mathbf{I}_t,\mathbf{s}_t\}$, where $\mathbf{I}_t$ is the visual input (e.g., multi-view RGB-D) and $\mathbf{s}_t$ is proprioception (e.g., joint angles, gripper states).
Given an instruction $\mathbf{l}$, a policy $\pi_\theta$ predicts the action
$\mathbf{a}_t=\pi_\theta(\mathbf{o}_t,\mathbf{l})$.
The model is trained via behavioral cloning (BC):
\begin{equation}
    \mathcal{L}_{\text{BC}}(\theta)
    = \mathbb{E}_{(\mathbf{o}_t,\mathbf{l},\mathbf{a}_t)\sim \mathcal{D}}
    \big[\ell(\pi_\theta(\mathbf{o}_t,\mathbf{l}), \mathbf{a}_t)\big],
\end{equation}
where $\ell(\cdot,\cdot)$ is an action-level loss (e.g., MSE or cross-entropy).

\paragraph{Multi-arm extension.}
For $N$ arms, arm $i$ has local state $\mathbf{s}_t^i$, local view $\mathbf{v}_t^i$, and action $\mathbf{a}_t^i$.
A joint policy predicts $\mathbf{A}_t=\{\mathbf{a}_t^1,\dots,\mathbf{a}_t^N\}$:
\begin{equation}
    \mathbf{A}_t
    = \Pi_\theta(\mathbf{I}_t, \mathbf{s}_t^1,\dots,\mathbf{s}_t^N,\mathbf{l}).
\end{equation}
We optimize the summed per-arm BC loss:
\begin{equation}
    \mathcal{L}_{\text{BC}}(\theta)
    = \mathbb{E}_{(\mathbf{O}_t,\mathbf{l},\mathbf{A}_t)\sim\mathcal{D}}
    \Big[\sum_{i=1}^N \ell\big(\pi_\theta^i(\mathbf{O}_t,\mathbf{l}), \mathbf{a}_t^i\big)\Big],
\end{equation}
where $\mathbf{O}_t$ aggregates per-arm observations and $\pi_\theta^i$ extracts arm-$i$ actions from $\Pi_\theta$.

\subsection{Multi-arm Compositional Generalization}

We study multi-arm compositional generalization: solving unseen collaborative executions by recombining the same atomic actions observed in training, without retraining as shown in Figure~\ref{fig:compositional_generalization}.

\paragraph{Atomic actions.}
Let $\mathcal{U}=\{u_1,\ldots,u_K\}$ denote a library of atomic actions (e.g., grasp, lift, align, place), shared across arms.

\paragraph{Composing a multi-arm execution.}
A multi-arm run can be written as a sequence of arm-tagged atomic actions
\begin{equation}
\mathbf{P}=\big[(i_1,u_{k_1}),\ldots,(i_T,u_{k_T})\big],
\end{equation}
where $(i_t,u_{k_t})$ means arm $i_t$ executes atomic action $u_{k_t}$ at step $t$ (possibly in parallel with other arms).

\paragraph{Generalization setting.}
Training covers compositions $\mathcal{P}_{\text{train}}$; at test time we require new compositions $\mathcal{P}_{\text{test}}$:
\begin{equation}
\mathcal{U}_{\text{test}}=\mathcal{U}_{\text{train}}, \qquad
\mathcal{P}_{\text{test}}\not\subset \mathcal{P}_{\text{train}}.
\end{equation}
Thus, the atomic actions remain in-distribution, while the collaboration structure is out-of-distribution.
Concretely, the shift comes from new (i) arm-role assignments ($i_t$), (ii) ordering and synchronization, and (iii) the resulting intermediate states and interactions.

\section{Method}
\label{sec:method}

\begin{figure*}[h!]
    \centering
    \includegraphics[width=1.0\linewidth]{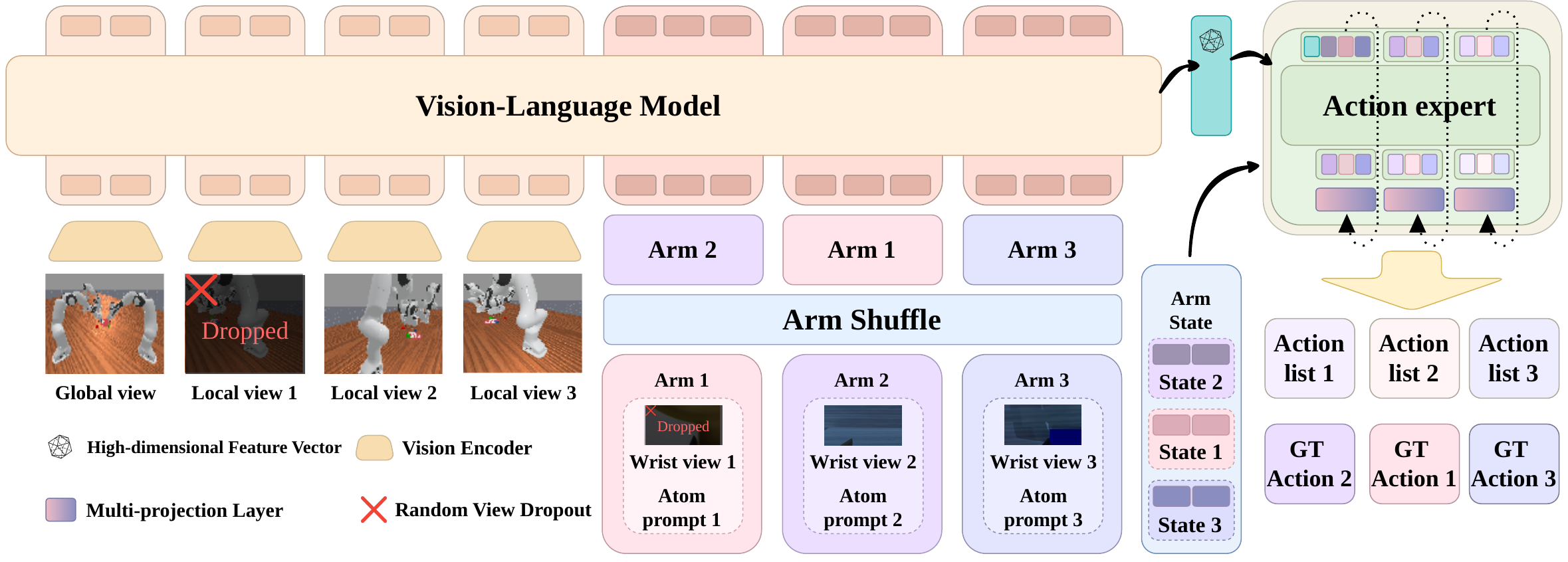}
    \caption{\textbf{MA-VLA Pipeline.}
    A high-level instruction is first decomposed into a sequence of atomic prompts.
Each arm is represented by a tuple containing its atomic prompt, state, wrist-view observation, and ground-truth action (e.g., \textbf{Arm1:}\raisebox{-0.3em}{\includegraphics[height=1em]{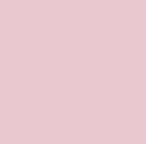}}, \textbf{Arm2:}\raisebox{-0.3em}{\includegraphics[height=1em]{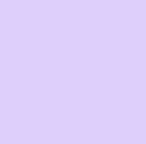}}, \textbf{Arm3:}~\raisebox{-0.3em}{\includegraphics[height=1em]{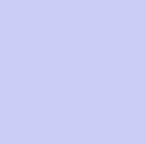}}).
During training, \textit{Arm Shuffle} randomly permutes these arm tuples, preventing the model from binding behaviors to fixed arm identities and promoting role-agnostic coordination across diverse collaboration patterns.}
    \label{fig:training_pipeline2}
\end{figure*}

\subsection{System Overview}

MA-VLA is a unified architecture for language-guided multi-arm coordination.  
It consists of two key components:  
(1) a \textbf{VLM-based Planner} that decomposes complex tasks into temporally ordered atomic sub-goals for each arm, and  
(2) a \textbf{VLA Executor} that grounds these linguistic arm-wise sub-goals into actions.

As shown in Figure~\ref{fig:training_pipeline}, during execution, the planner interprets a high-level instruction $\mathbf{l}$ and visual observations $\mathbf{I}_t$, producing atomic prompts $\{\mathbf{p}_1, \dots, \mathbf{p}_T\}$ that specify each arm's behavior at each stage.  
The VLA executor then conditions on the current scene and state to generate continuous control actions for each arm.  
This hierarchical language interface lets MA-VLA jointly reason about collaboration structure (planner) and fine-grained physical control (executor).  
A flow-matching expert~\cite{lipman2022flow} further ensures temporally smooth, physically consistent execution across all arms.

\subsection{VLM-based Planner: Task Decomposition via Atomic Prompts}

To mirror how humans coordinate through language, we introduce a VLM-based planner $\mathcal{P}_{\phi}$ that converts a high-level goal into a structured sequence of per-arm sub-instructions.
Instead of mapping $\mathbf{l}$ directly to low-level control, the planner predicts a stage-wise assignment of atomic prompts by jointly conditioning on the instruction $\mathbf{l}$, visual observations $\mathcal{I}$, and a predefined atomic prompt set $\mathcal{A}$.
Here $\mathcal{I}$ denotes a set of images available at planning time, which can come from multi-view observations or consecutive frames (e.g., from a short video clip), without assuming a specific frame-sampling procedure.
This yields an interpretable interface for multi-arm coordination while keeping the downstream executor fully end-to-end.

Formally, the instruction $\mathbf{l}$ is decomposed into $T$ atomic stages:
\begin{equation}
\mathbf{l} = (\mathbf{p}_1,\dots,\mathbf{p}_T),\qquad
\mathbf{p}_t = (p_t^1,\dots,p_t^N),
\end{equation}
where $p_t^i \in \mathcal{A}$ denotes the atomic prompt assigned to arm $i$ at stage $t$ (e.g., \texttt{grasp object}, \texttt{hold bowl}, \texttt{place cube}).
We use stages to represent temporally grounded sub-goals, each executed until a task-dependent termination condition is met (Appendix).

Given $\mathcal{I}$ and $\mathbf{l}$, the planner generates the full prompt sequence in a single forward pass:
\begin{equation}
(\mathbf{p}1,\dots,\mathbf{p}T) = \mathcal{P}{\phi}(\mathcal{I}, \mathbf{l}, \mathcal{A}).
\end{equation}
Here, $\mathcal{A}$ is a curated and finite set of fine-grained prompt templates shared across tasks, which defines a reproducible action vocabulary (the full list is provided in the supplementary material).
In practice, we prompt the VLM to select templates from $\mathcal{A}$ and canonicalize its outputs to valid templates in $\mathcal{A}$, ensuring that each $p_t^i$ corresponds to a well-defined atomic instruction.
By reasoning over spatial and semantic cues in $\mathcal{I}$, $\mathcal{P}{\phi}$ produces temporally coherent and semantically grounded sub-goals that explicitly specify who performs which action at each stage.
The planner outputs only language prompts and is not fine-tuned; all control actions are produced by the learned executor.
Implementation details, the prompting format, and hyperparameters are provided in the supplementary material.

\subsection{VLA Executor: Multi-arm Action Generation}

Given the set of atomic prompts produced by the planner, the VLA executor $\pi_\theta$ grounds these linguistic sub-goals into executable control actions through multimodal reasoning.
Rather than predicting each arm’s motion independently, MA-VLA adopts a unified model that jointly infers all arms’ actions within a single forward pass, enabling both coordinated and individualized behaviors.

At each timestep $t$, the atomic prompts for all arms are concatenated into a unified instruction string:
\begin{equation}
\mathbf{u}_t = \text{Arm0: } p_t^0 \text{, Arm1: } p_t^1 \text{, ... , ArmN: } p_t^N,
\end{equation}
where $\mathbf{u}_t$ represents the aggregated linguistic context describing the collaborative intent of all arms.
This unified prompt serves as the language input to the VLA executor.

The executor then takes as input: (1) multi-view visual observations $\mathbf{I}_t = {\mathbf{I}_t^{view}, \mathbf{I}_t^{wrist}}$ that include both global and egocentric perspectives, (2) the concatenated proprioceptive states of all arms $\mathbf{s}_t = [\mathbf{s}_t^1, \dots, \mathbf{s}_t^N]$, and (3) the unified instruction $\mathbf{u}_t$.
The VLA executor outputs the complete multi-arm action vector:
\begin{equation}
[\mathbf{a}_t^1, \dots, \mathbf{a}_t^N] = \pi\theta(\mathbf{I}_t, \mathbf{s}_t, \mathbf{u}_t).
\end{equation}

By conditioning on the shared visual-linguistic context $\mathbf{u}_t$, MA-VLA effectively captures inter-arm dependencies and ensures coherent coordination among arms, while preserving the flexibility for each arm to perform distinct yet complementary actions.

To produce smooth and physically consistent control trajectories, we adopt a lightweight flow-matching formulation inspired by pi0~\cite{black2024pi_0}. During training, the VLA predicts a denoising vector field that maps noisy latent actions $\mathbf{x}_t^{\tau}$ toward the ground-truth actions $\mathbf{a}_t$, conditioned on visual observations $\mathbf{I}_t$, proprioceptive states $\mathbf{s}_t$, and the unified linguistic context $\mathbf{u}_t$. At inference time, actions are obtained by integrating the learned flow field from noise to clean latent space. To better support multi-arm collaboration, we further include a multi-head projection layer that separates the shared latent representation into arm-specific action heads.

\subsection{Training Strategies for Compositional Generalization}

Although MA-VLA learns to execute atomic actions via multimodal conditioning, we find that generalization remains limited when arms are tied to fixed spatial roles or when perception depends heavily on arm-specific states (e.g., visual inputs and proprioceptive states).
To address these limitations, we introduce two stochastic regularization strategies that are applied only during training: \textbf{Arm Shuffle} and \textbf{View Dropout}.
Both operate as data-level perturbations that promote role-invariant reasoning and redundancy-aware perception, while leaving the training objective unchanged.

\paragraph{Arm Shuffle.}
In each training iteration, with a probability $p_{\text{shuffle}}$, we randomly permute the correspondence between arms’ states, views, prompts, and actions:
\begin{equation}
    (\mathbf{s}_t^i, \mathbf{v}_t^i, \mathbf{p}_t^i, \mathbf{a}_t^i)
    \xrightarrow{\text{shuffle}}
    (\mathbf{s}_t^{\sigma(i)}, \mathbf{v}_t^{\sigma(i)}, \mathbf{p}_t^{\sigma(i)}, \mathbf{a}_t^{\sigma(i)}),
\end{equation}
where $\sigma \in \mathcal{S}_N$ is a random permutation of the $N$ arm indices.  
This stochastic role reassignment prevents the model from overfitting to fixed positional or identity-based correlations, forcing each arm to interpret its atomic prompt semantically rather than structurally.

\paragraph{View Dropout.}
To enhance perception robustness, we apply random masking to a subset of visual inputs with probability $p_{\text{drop}}$.  
Let $\mathcal{M}\subseteq\{1,\dots,M\}$ denote the indices of dropped camera views.  
We define the masked observation as
\begin{equation}
    \tilde{\mathbf{I}}_t = \text{Mask}(\mathbf{I}_t, \mathcal{M}),
\end{equation}
where $\text{Mask}(\cdot)$ replaces selected views with zeros.  
This encourages the model to utilize redundant spatial cues and learn consistent multi-view reasoning.

\paragraph{Training objective.}
Both strategies are integrated as stochastic augmentations under a unified behavioral cloning objective.  
At each iteration, the model samples whether to apply shuffle or dropout, and computes the same action-level loss:
% \begin{equation}
%     \mathcal{L}_{\text{train}}(\theta)
%     = \mathbb{E}_{(\mathbf{O}_t,\mathbf{l},\mathbf{A}_t)\sim\mathcal{D}}
%       \Bigg[
%         \sum_{i=1}^{N}
%         \ell\big(
%           \pi_\theta(\tilde{\mathbf{I}}_t, \mathbf{s}_t^{i}, \mathbf{p}_t^{i}),
%           \mathbf{a}_t^{i}
%         \big)
%       \Bigg],
% \end{equation}
\begin{equation}
\mathcal{L}_{\text{t}}(\theta)
= \mathbb{E}_{(\mathbf{O}_t,\mathbf{l},\mathbf{A}_t)\sim\mathcal{D}}
\left[
\sum_{i=1}^{N}
\ell\!\left(
\pi_\theta(\tilde{\mathbf{I}}_t,\mathbf{s}_t^{i},\mathbf{p}_t^{i}),
\mathbf{a}_t^{i}
\right)
\right].
\end{equation}

where $\tilde{\mathbf{I}}_t$ and ${\mathbf{s}_t^i, \mathbf{p}t^i, \mathbf{a}t^i}$ denote the possibly perturbed inputs after applying stochastic \emph{Arm Shuffle} and \emph{View Dropout}, controlled respectively by the shuffle rate $p_{\text{shuffle}}$ and the view-drop rate $p_{\text{drop}}$.
These perturbations preserve the original learning objective while implicitly regularizing MA-VLA toward permutation-invariant coordination and more robust visual grounding.
% where $\tilde{\mathbf{I}}_t$ and $\{\mathbf{s}_t^i, \mathbf{p}_t^i, \mathbf{a}_t^i\}$ may have been modified by the stochastic shuffle or dropout process according to their respective probabilities $p_{\text{shuffle}}$ and $p_{\text{drop}}$.  
% This preserves the learning objective while implicitly regularizing MA-VLA toward permutation-invariant coordination and robust visual grounding.
\section{Experiments}
\label{sec:experiments}

In this section, we evaluate MA-VLA under both in-domain settings and multi-arm compositional generalization (out-of-domain) splits.

\subsection{Experimental Setup}
\label{sec:exp_setup}

\paragraph{Training configurations.}
We report two training configurations.
\textbf{Clean} disables Arm Shuffle and View Dropout, training on the original demonstrations.
\textbf{Regularized} enables Arm Shuffle and View Dropout during training.

\paragraph{Benchmarks.}
We evaluate two simulation benchmarks and one real-robot platform, reporting in-domain performance and out-of-domain compositional generalization.
\textbf{RoboFactory}\cite{qin2025robofactory} contains cooperative manipulation tasks with 2–4 arms, emphasizing coordination and parallel execution.
\textbf{RoboTwin2.0}\cite{chen2025robotwin} focuses on dual-arm collaboration under stronger visual disturbances, including additional distractors, background variation, and lighting changes; we use standard multi-view observations with wrist-camera views.
\textbf{Real-world SO101} is a dual-arm system with 12 degrees of freedom (6 per arm), equipped with one global RGB camera and one wrist camera per arm; all cameras run at 30Hz. Real-robot tasks include Stack Bowls, Place Cubes, Pass Toys, and Stack Cubes; details are in the supplementary material.

\paragraph{Demonstrations and atomic action labels.}
For each simulation task, we collect 150 expert demonstrations in the in-domain setting.
Each trajectory contains synchronized multi-view observations, per-arm proprioception, and action sequences.
We convert demonstrations into frame-level atomic action prompts using a rule-based parser with task-specific state predicates (e.g., contact, grasp status, object pose thresholds), producing training tuples $(\mathbf{O}_t,\mathbf{l}_t,\mathbf{A}_t)$ where $\mathbf{l}_t$ is the atomic prompt at time $t$.

\paragraph{Evaluation protocol and metrics.}
We evaluate two settings.
\textbf{In-domain Collaboration} tests the same task distribution as training with randomized perturbations (object pose shifts, camera viewpoint changes, and environment variations).
\textbf{Out-of-domain Compositional Generalization} tests unseen collaboration structures where atomic actions remain in-distribution but are recomposed into unseen temporal orders, role assignments, or interaction patterns.
For simulation, each configuration is evaluated over 100 rollouts and we report mean success rate; success is defined by task-specific goal predicates in the simulator.
Unless otherwise stated, we use the same perturbation suite across methods. For compositional generalization splits, we enable Arm Shuffle and View Dropout during training; otherwise we train on the original demonstrations.

\paragraph{Real-world data and evaluation.}
On SO101, we evaluate four bimanual tasks: Stack Two Bowls, Place Two Cubes, Pass Two Toys, and Stack Two Cubes. The detailed real-robot task settings are provided in the supplementary material. For each task, we collect 50 teleoperated demonstrations using the LeRobot~\cite{cadene2026lerobot}, train for 15{,}000 gradient steps with a batch size of 32, and evaluate over 20 episodes. We report the success rate using the same in-domain and out-of-domain split definition as in simulation. Each trial starts from a randomized but valid initial configuration and stops upon success, timeout, or a safety stop.

\paragraph{Backbones and baselines.}
Our framework uses Pi0~\cite{black2024pi_0} initialized from the official \texttt{pi0\_base} checkpoint.
The VLM planner uses GPT-4.1 to generate atomic action prompts (see supplementary material for the prompt template and decoding settings).
We compare against representative baselines:
ACT~\cite{zhao2023learning}, DP~\cite{chi2025diffusion}, DP3~\cite{ze20243d}, and Pi0-FAST~\cite{pertsch2025fast}.
Since prior VLA policies are designed for single-arm inputs, we adapt their input/output interfaces to multi-arm RoboFactory and retrain them from the same demonstrations with matched training budgets.

\subsection{Implementation Details}
\label{sec:impl}

All experiments use two NVIDIA A800 GPUs with batch size 32.
For RoboFactory, we train 2-arm tasks for 10{,}000 steps and 3--4 arm tasks for 15{,}000 steps.
For RoboTwin~2.0 (Hard), we train for 30{,}000 steps.
The horizon is 50 for all tasks.
For the compositional setting, we set Arm Shuffle probability and View Dropout probability per out-of-domain task:
Three Robots Stack Cube (Reordering): $p_{\text{shuffle}}{=}0.8$, $p_{\text{drop}}{=}0.2$;
Pass Two Shoes (Extended Collaboration): $p_{\text{shuffle}}{=}1.0$, $p_{\text{drop}}{=}0.4$;
Reverse Stack Bowls (Reordering+Disturbance): $p_{\text{shuffle}}{=}0.6$, $p_{\text{drop}}{=}0.4$. For SO101, we enable Arm Shuffle throughout training to reduce arm-identity bias and encourage instruction-centric coordination.
Specifically, we set $p_{\text{shuffle}}{=}1.0$ and apply View Dropout (masking) with probability $p_{\text{drop}}{=}0.4$ on multi-camera observations.
All other training hyperparameters follow the simulation setup unless otherwise stated.
We keep all other hyperparameters identical across methods within each benchmark.

\begin{table}[t]
\centering
\caption{
Performance on RoboFactory Benchmark~\cite{qin2025robofactory}.
Abbrev.: Lift=Lift Barrier, Place=Place Food, Stack2=Two Arms Stack Cube, Pass=Pass Shoe;
Stack3=Three Arms Stack Cube, Align=Camera Alignment, Deliver=Long Pipeline Delivery, Photo=Take Photo.
}
\label{tab:collaboration_robofactory}
\footnotesize
\setlength{\tabcolsep}{4pt}
\renewcommand{\arraystretch}{1.12}

\begin{tabularx}{\columnwidth}{l *{5}{>{\centering\arraybackslash}X}}
\toprule
\multicolumn{6}{c}{Two-arm tasks} \\
\midrule
Method & Lift & Place & Stack2 & Pass & Avg \\
\midrule
DP~\cite{chi2025diffusion}              & 58.0\% & 20.0\% & 20.0\% & 12.0\% & 27.5\% \\
Pi0-Fast~\cite{pertsch2025fast}         & 90.0\% & 27.0\% & 50.0\% & 63.0\% & 57.5\% \\
Pi0~\cite{black2024pi_0}                & 97.0\% & 58.0\% & 82.0\% & 84.0\% & 80.3\% \\
\rowcolor{gray!12}
MA-VLA                                  & \textbf{98.0\%} & \textbf{59.0\%} & \textbf{86.0\%} & \textbf{91.0\%} & \textbf{83.5\%} \\
\midrule
\cmidrule(lr){2-3}\cmidrule(lr){4-5}
& \multicolumn{2}{c}{Three-arm Tasks} & \multicolumn{2}{c}{Four-arm Tasks} & \\
\midrule
Method & Stack3 & Align & Deliver & Photo & Avg \\
\midrule
DP~\cite{chi2025diffusion}              & 22.0\% & 19.0\% & 0.0\%  & 20.0\% & 15.3\% \\
Pi0-Fast~\cite{pertsch2025fast}         & 3.0\%  & 22.0\% & 10.0\% & 12.0\% & 11.8\% \\
Pi0~\cite{black2024pi_0}                & 48.0\% & 94.0\% & 82.0\% & \textbf{82.0\%} & 76.5\% \\
\rowcolor{gray!12}
MA-VLA                                  & \textbf{58.0\%} & \textbf{96.0\%} & \textbf{97.0\%} & \textbf{82.0\%} & \textbf{83.3\%} \\
\bottomrule
\end{tabularx}
\end{table}

\begin{table}[h!]
\centering
\caption{
Performance on RoboTwin 2.0 (Hard)~\cite{chen2025robotwin}.
Abbrev.: Shoes=Place Dual Shoes; Handover=Handover Block; Bowls=Stack Bowls Two;
Roller=Grab Roller; Cabinet=Place Object (Cabinet); Skillet=Place Bread (Skillet); Stamp=Stamp Seal.
}
\label{tab:collaboration_robotwin}
\setlength{\tabcolsep}{3.2pt}
\renewcommand{\arraystretch}{1.08}
\footnotesize
\resizebox{\linewidth}{!}{%
\begin{tabular}{lcccccccc}
\toprule
Method & Shoes & Handover & Bowls & Roller & Cabinet & Skillet & Stamp & Avg \\
\midrule
ACT~\cite{zhao2023learning} & 0.0 & 0.0 & 0.0 & 58.0 & 23.0 & 5.0 & 0.0 & 12.3 \\
DP~\cite{chi2025diffusion} & 4.0 & 21.0 & 1.0 & 58.0 & 42.0 & 6.0 & 0.0 & 18.9 \\
DP3~\cite{ze20243d} & 7.0 & 14.0 & 39.0 & 78.0 & 52.0 & 9.0 & 1.0 & 28.6 \\
Pi0-FAST~\cite{pertsch2025fast} & 10.0 & 25.0 & 44.0 & 83.0 & 23.0 & 2.0 & 2.0 & 27.0 \\
Pi0~\cite{black2024pi_0} & 14.0 & 27.0 & 59.0 & 98.0 & 59.0 & 28.0 & 3.0 & 41.1 \\
\rowcolor{gray!12}
MA-VLA & \textbf{22.0} & \textbf{46.0} & \textbf{63.0} & \textbf{99.0}
& \textbf{71.0} & \textbf{30.0} & \textbf{12.0} & \textbf{49.0} \\
\bottomrule
\end{tabular}%
}
\end{table}

\subsection{In-domain Collaboration Results}
\label{sec:in_domain}

Table~\ref{tab:collaboration_robofactory} shows that conditioning on atomic action instructions consistently improves in-domain collaboration on RoboFactory.
The gains grow with more arms (3–4 arms), suggesting atomic prompts better disambiguate per-arm responsibilities as coordination complexity increases.
On RoboTwin2.0 (Hard), Table~\ref{tab:collaboration_robotwin} shows MA-VLA remains robust under strong visual disturbances, consistently improving over the Pi0 backbone and diffusion-based baselines.

\subsection{Out-of-domain Compositional Generalization Results}
\label{sec:out_domain}

We evaluate compositional generalization across unseen collaboration patterns in RoboFactory and RoboTwin2.0 (Hard).
As shown in Table~\ref{tab:moreagent_compositional}, MA-VLA achieves non-trivial success where end-to-end imitation baselines collapse, improving out-of-domain success by up to 13.0% over Pi0 and Pi0-FAST.
These scenarios contain novel role orders and coordination structures unseen in training.
Overall, atomic action conditioning, Arm Shuffle, and View Dropout reduce reliance on fixed arm identities and promote instruction-centric coordination, enabling recomposition of learned atomic skills under unseen collaboration paradigms.

\begin{table}[h!]
\centering
\caption{
Results on unseen out-of-domain collaboration tasks. Arrows denote the required stacking order of cube colors.
Notation: GBR=G$\!\to$B$\!\to$R; RGB=R$\!\to$G$\!\to$B; BRG=B$\!\to$R$\!\to$G.
}
\label{tab:moreagent_compositional}
\setlength{\tabcolsep}{3.2pt}
\renewcommand{\arraystretch}{1.08}
\footnotesize
\resizebox{\linewidth}{!}{%
\begin{tabular}{lcccccc}
\toprule
Method & GBR & RGB & BRG & Pass Two Shoes & Stack Two Bowls & Avg \\
\midrule
DP~\cite{chi2025diffusion}      & 0.0 & 0.0 & 0.0 & 0.0 & 0.0 & 0.0 \\
Pi0-FAST~\cite{pertsch2025fast} & 0.0 & 0.0 & 0.0 & 0.0 & 0.0 & 0.0 \\
Pi0~\cite{black2024pi_0}        & 0.0 & 0.0 & 0.0 & 0.0 & 0.0 & 0.0 \\
\rowcolor{gray!12}
MA-VLA                          & \textbf{28.0} & \textbf{9.0} & \textbf{9.0} & \textbf{9.0} & \textbf{10.0} & \textbf{13.0} \\
\bottomrule
\end{tabular}%
}
\end{table}

% We evaluate compositional generalization on unseen collaboration patterns across RoboFactory and RoboTwin~2.0 (Hard).
% As shown in Table~\ref{tab:moreagent_compositional}, MA-VLA achieves non-trivial success in settings where standard end-to-end imitation baselines often collapse, improving out-of-domain success by up to 13.0\% over strong backbones such as Pi0 and Pi0-FAST.
% These test scenarios involve novel role orders and coordination structures that never appear during training.
% Overall, atomic action conditioning combined with Arm Shuffle and View Dropout reduces reliance on fixed arm identities and encourages instruction-centric coordination, enabling recomposition of learned atomic skills under unseen collaboration paradigms.

\subsection{Real-world Dual SO101 Results}
\label{sec:real_world}

Table~\ref{tab:so101_real} summarizes real-world results.
MA-VLA consistently improves Pi0 in-domain, and importantly achieves non-zero success in role-reversed out-of-domain settings where Pi0 fails.
This suggests that atomic action conditioning, together with Arm Shuffle and View Dropout, mitigates arm-identity bias and improves transfer to unseen real-world collaboration patterns.

\begin{table}[t]
\centering
\caption{Real-world results on dual-arm SO101.
Each task is evaluated in-domain (ID) and compositional out-of-domain (OOD).
OOD reconfigures the program: Stack Bowls/Stack Cubes use reversed stacking, Place Cubes uses reversed placement, and Pass Toys uses an alternative passing order.}
\label{tab:so101_real}
\footnotesize
\setlength{\tabcolsep}{3.8pt}
\renewcommand{\arraystretch}{1.12}
\begin{tabularx}{\columnwidth}{l *{4}{>{\centering\arraybackslash}X X}}
\toprule
& \multicolumn{2}{c}{Stack Bowls}
& \multicolumn{2}{c}{Place Cubes}
& \multicolumn{2}{c}{Pass Toys}
& \multicolumn{2}{c}{Stack Cubes} \\
\cmidrule(lr){2-3}\cmidrule(lr){4-5}\cmidrule(lr){6-7}\cmidrule(lr){8-9}
Method
& ID & OOD
& ID & OOD
& ID & OOD
& ID & OOD \\
\midrule
Pi0~\cite{black2024pi_0}
& 9/20  & 0/20
& 12/20 & 0/20
& 3/20  & 0/20
& 5/20  & 0/20 \\
\rowcolor{gray!12}
MA-VLA
& \textbf{12/20} & \textbf{10/20}
& \textbf{15/20} & \textbf{8/20}
& \textbf{6/20}  & \textbf{2/20}
& \textbf{8/20}  & \textbf{2/20} \\
\bottomrule
\end{tabularx}
\end{table}

\begin{figure}[t]
    \centering
    \includegraphics[width=1.0\linewidth]{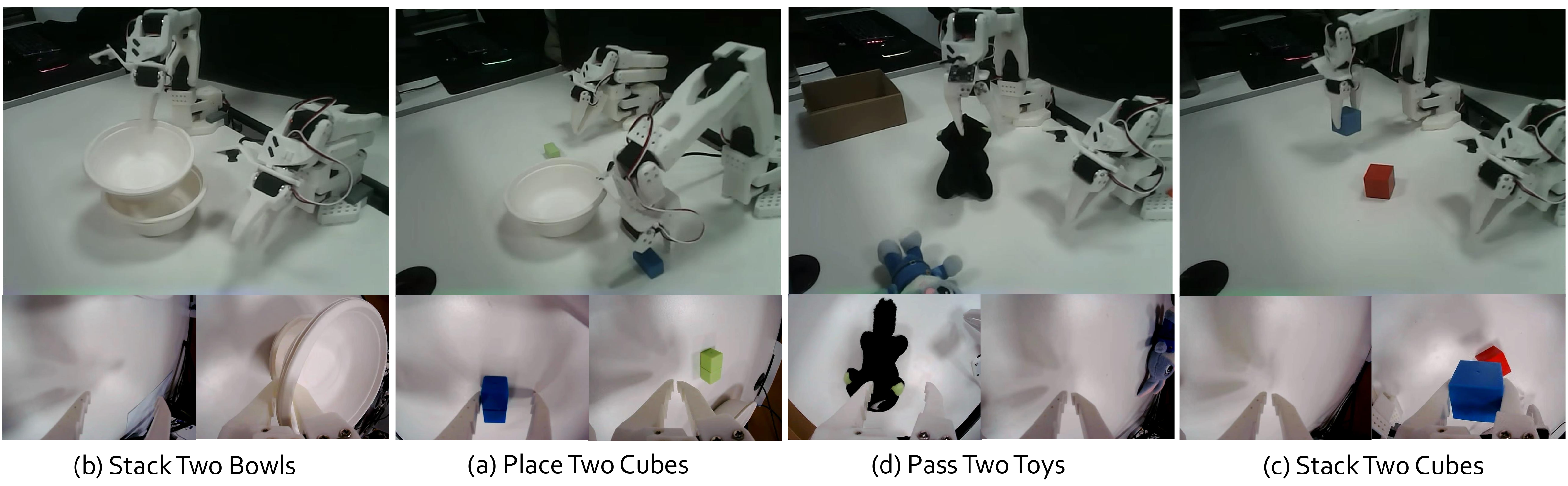}
    \caption{Setup of the real-world dual-arm SO101 tasks. Top: global view. Bottom: wrist views (left: left-wrist camera; right: right-wrist camera).}
    \label{fig:example1}
\end{figure}

\section{Ablation Study}
We use the Three Robots Stack Cube unseen-order task for evaluation. The in-domain order is blue–green–red, while the out-of-domain configurations include green–blue–red (G$! \to$B$! \to$R), red–green–blue (R$! \to$G$! \to$B), and blue–red–green (B$! \to$R$! \to$G). For out-of-domain results, we average performance across the three unseen configurations.

\subsection{Ablation of Components}
We ablate our components on the Three Robots Stack Cube unseen-order task. As shown in Table~\ref{ab:component}, introducing atomic actions markedly improves in-domain performance. Adding Arm Shuffle yields a breakthrough in out-of-domain generalization, achieving non-zero success where prior models fail and demonstrating clearer compositional generalization. Since generalization involves substantial visual discrepancies (e.g., environment changes, target appearance, and arm configurations), we also incorporate View Dropout. This further boosts out-of-domain compositional generalization with only minimal in-domain degradation.

\begin{table}[h!]
\centering
\caption{Ablation Experiment of Components.}
\label{ab:component}
\resizebox{0.6\linewidth}{!}{
\begin{tabular}{cccccc}
\toprule
Atom & Shuffle & View Dropout & Out-of-domain & In-domain \\
\midrule
 &  &    & 0.0\% & 48.0\%  \\
\checkmark &  &  & 0.0\% & \textbf{58.0\%} \\
\checkmark &   \checkmark &  & 7.3\% & 52.0\% \\
\checkmark &  \checkmark & \checkmark & \textbf{15.3\%} & 53.0\% \\
\bottomrule
\end{tabular}
}
\end{table}

\subsection{Ablation of Shuffle Probability}
\begin{figure}[t]
    \centering
    \includegraphics[width=0.60\linewidth]{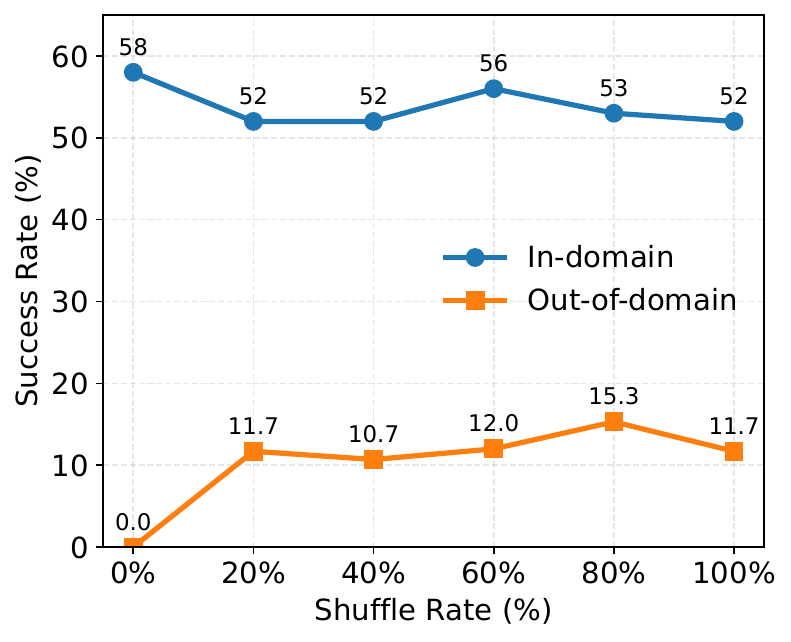}
    \caption{Effect of shuffle rate on in- and out-of-domain settings.}
    \label{fig:training_pipeline}
\end{figure}
We further analyze the impact of the shuffle rate through a controlled ablation study, where $p_{\text{shuffle}}$ determines the probability of randomly permuting arms’ states, viewpoints, and instructions. As shown in Figure~\ref{fig:training_pipeline}, we vary the shuffle probability from 0\% to 100\%. Increasing the shuffle rate boosts out-of-domain performance—from near-zero to significantly higher levels—highlighting its strong effect on compositional generalization. While in-domain accuracy experiences a mild drop, it remains stable. This indicates that shuffle enables the model to gradually generalize better to new collaboration tasks.

\subsection{Ablation of Separate Model}
For multi-arm control, a simple baseline trains an independent VLA per arm. To build this setting, we split multi-arm trajectories into per-arm streams so each model sees only its own viewpoint, state, and action history, and we annotate arm-specific atomic actions for single-arm instruction following. At test time, we run three separate VLA models in parallel, one per arm. Results in Table~\ref{ab:sepmodel} show poor compositional generalization: while each arm follows its atomic actions, the models fail to coordinate on unseen multi-arm tasks. This design also introduces inference and deployment overhead that grows linearly with the number of arms, underscoring the limitations of independent-arm training and the scalability of our unified framework.

\begin{table}[h!]
\centering
\caption{Comparison between MA-VLA and Separate Model.}
\label{ab:sepmodel}
\resizebox{0.7\linewidth}{!}{
\begin{tabular}{lccccc}
\toprule
Method & Out-of-domain &  In-domain & Avg. &  VLA Number \\
\midrule
Pi0 & 0.0\% &  48.0\%& 24.0\% & \textbf{1} \\
Separate Pi0 & 0.0\% &  \textbf{61.0\%} & 30.5\% & 3 \\
MA-VLA & \textbf{15.3\%} &  53.0\%& \textbf{34.2\%} &\textbf{1} \\

\bottomrule
\end{tabular}
}
\end{table}

\section{Visualization}
Figure~\ref{fig:example1} shows MA-VLA rollouts in RoboFactory (simulation) and on the real-world SO101 under both in-domain and out-of-domain splits.
The out-of-domain split targets \emph{multi-arm compositional generalization}: the model must recombine seen atomic actions into unseen multi-arm coordination patterns.

\begin{figure}[h!]
    \centering
    \includegraphics[width=1.0\linewidth]{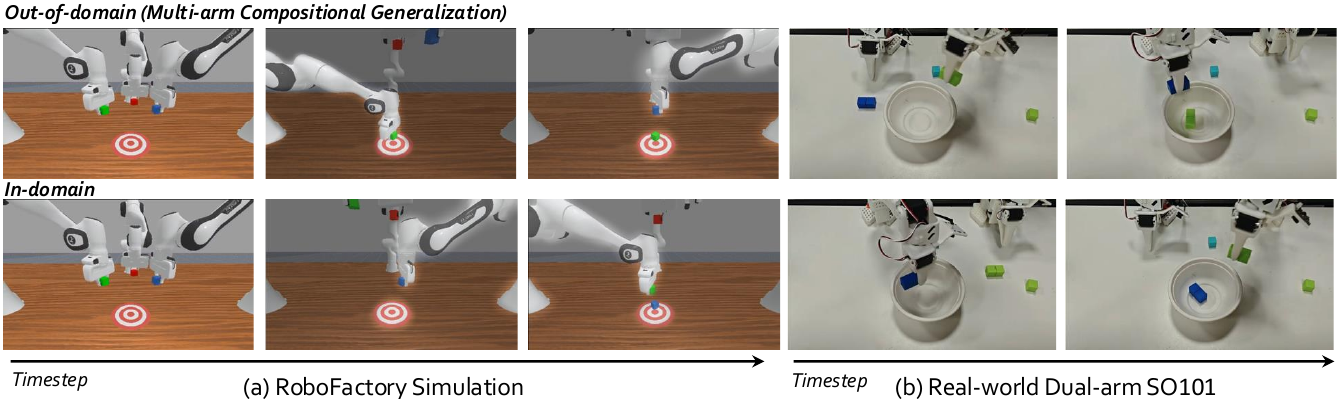}
    \caption{Qualitative rollouts of MA-VLA in RoboFactory and on SO101 under in-domain and out-of-domain splits. Out-of-domain evaluates compositional generalization by requiring unseen recombinations of atomic actions under novel arm and object states.}
    \label{fig:example1}
\end{figure}

\section{Conclusion}
We introduce MA-VLA, a unified VLA framework that makes multi-arm collaboration explicit by decomposing a high-level instruction into interpretable atomic actions and assigning them to individual arms within a single model. With Arm Shuffle to encourage role-agnostic behavior, MA-VLA improves compositional transfer to unseen collaboration patterns. Experiments on RoboFactory, RoboTwin 2.0, and real-world SO101 show consistent gains over strong imitation learning and VLA baselines in both in-domain performance and multi-arm compositional generalization, supporting a practical route to scalable multi-arm instruction following.

\section{Acknowledgment}
This paper is supported by the National Natural Science Foundation of China (U23A20386, 62422610, 62441231, 62276045, 62576072), Liao Ning Science and Technology Plan (2025JH2/101330121, 2025JH2/101330124, 2023JH26/10200016), and Dalian City Science and Technology Innovation Fund (2023JJ11CG001).

\bibliographystyle{splncs04}
\bibliography{main}
\end{document}